\documentclass{article} % For LaTeX2e
\usepackage{iclr2021_conference,times}

\usepackage{amsmath,amsfonts,bm}

\def\eqref#1{equation~\ref{#1}}
\def\1{\bm{1}}

\DeclareMathAlphabet{\mathsfit}{\encodingdefault}{\sfdefault}{m}{sl}
\SetMathAlphabet{\mathsfit}{bold}{\encodingdefault}{\sfdefault}{bx}{n}

\newcommand{\E}[2]{\mathbb{E}_{#1}{\left[#2\right]}}

\usepackage{hyperref}
\usepackage{url}
\usepackage{graphicx}
\usepackage{wrapfig, blindtext}
\usepackage{soul}
\usepackage{booktabs}
\usepackage{algorithm,algpseudocode}

\usepackage{amsmath,amsfonts,amssymb,amsthm, bm}
\usepackage{subcaption}
\title{Lyapunov Barrier Policy Optimization}

\iclrfinalcopy
\author{Harshit Sikchi  \\
Computer Science Department\\
Carnegie Mellon University\\
Pittsburgh, PA 15213, USA \\
\texttt{\{hsikchi\}@cs.cmu.edu} \\
\And
Wenxuan Zhou, David Held \\
Robotics Institute \\
Carnegie Mellon University \\
Pittsburgh, PA 15213, USA  \\
\texttt{\{wenxuanz,dheld\}@cs.cmu.edu} \\
}

\algnewcommand{\algorithmicforeach}{\textbf{for}}
\algdef{SE}[FOR]{ForEach}{EndForEach}[1]
  {\algorithmicforeach\ #1\ \algorithmicdo}% \ForEach{#1}
  {\algorithmicend\ \algorithmicforeach}% \EndForEach

\begin{document}
\setlength\intextsep{0pt}

\maketitle

\begin{abstract}
Deploying Reinforcement Learning (RL) agents in the real-world require that the agents satisfy safety constraints. Current RL agents explore the environment without considering  these constraints, which can lead to damage to the hardware or even other agents in the environment. We propose a new method, LBPO, that uses a Lyapunov-based barrier function to restrict the policy update to a safe set for each training iteration. 
Our method also allows the user to control the conservativeness of the agent with respect to the constraints in the environment. 
LBPO significantly outperforms state-of-the-art baselines in terms of the number of constraint violations during training while being competitive in terms of performance.  Further, our analysis reveals that baselines like CPO and SDDPG rely mostly on backtracking to ensure safety rather than safe projection, which provides insight into why previous methods might not have effectively limit the number of constraint violations.
\end{abstract}

\section{Introduction}

Current reinforcement learning methods are trained without any notion of safe behavior. As a result, these methods might cause damage to themselves, their environment, or even harm other agents in the scene. Ideally, an agent in a real-world setting should start with a conservative policy and iteratively refine it while maintaining safety constraints. 
% For example, a human learning to play the game of table-tennis will ensure that his racket does not hit the table as he/she tries to figure out the game, a person trying to learn a car will  ensure that he does not crash into pedestrians and other vehicles. 
For example, an agent that is learning to drive around other agents should start driving slowly and gradually learn to improve its performance by exploring carefully while avoiding accidents. In contrast, most deep reinforcement learning methods 
learn by trial and error without taking into consideration the safety-related consequences of their actions~\citep{silver2016mastering,vinyals2019grandmaster,akkaya2019solving}.  In this work, we address the problem of learning control policies that optimize a reward function while satisfying some predefined constraints throughout the learning process. 

% has shown promise in a number of complicated domains like Go~\citep{silver2016mastering}, Starcraft~\citep{vinyals2019grandmaster} and Dexterous Rubiks Cube Manipulations~\citep{akkaya2019solving}. These algorithms 

As in previous work for safe reinforcement learning, we use human-defined constraints to specify safe behavior. A classical model for RL with constraints is the constrained Markov Decision Process (CMDP)~\citep{altman1999constrained}, where an agent tries to maximize the standard RL objective of expected returns while satisfying constraints on expected costs. A number of previous works on CMDPs mainly focus on environments that have low dimensional action spaces, and they are difficult to scale up to more complex environments~\citep{turchetta2019safe, wachi2020safe}. One popular approach to solve Constrained MDPs in large state and action spaces is to use the Lagrangian method~\citep{altman1998constrained, Ray2019}. This method augments the original RL objective with a penalty on constraint violations and computes the saddle point of the constrained policy optimization via primal-dual methods~\citep{altman1998constrained}. While safety is ensured asymptotically,  no guarantees are made about safety during training. As we show, other methods which claim to maintain safety during training also lead to many safety violations during training in practice. In other recent work, \cite{chow2018lyapunov, chow2019lyapunov} use Lyapunov functions to explicitly model constraints in the CMDP framework to guarantee safe policy updates. We build on this idea, where the Lyapunov function allow us to convert trajectory-based constraints in the CMDP framework to state-based constraints which are much easier to deal with.

% Lyapunov functions have been commonly used in control theory to study dynamical stability~\citep{perkins2002lyapunov}. 
% A Lyapunov function is a scalar potential function that keeps track of the energy a system continuously dissipates. \citep{berkenkamp2017safe} used Lyapunov functions to guarantee exploration in a model based RL  such that it can also return back to its region of attraction. 

In this work, we present a new method called LBPO (Lyapunov Barrier Policy Optimization) for safe reinforcement learning in the CMDP framework.  We formulate the policy update as an unconstrained update augmented by a barrier function which ensures that the policy lies in the set of policies induced by the Lyapunov function, thereby guaranteeing safety. We show that LBPO allows us to control the amount of risk-aversion of the agent by adjusting the barrier. We also analyze previous baselines that use a backtracking recovery rule and empirically show that their near-constraint satisfaction can be explained by their recovery rule; this approach leads to many constraint violations in practice. Finally, we demonstrate that LBPO outperforms state-of-the-art CMDP baselines in terms of the number of constraint violations while being competitive in performance.

\section{Background}

We consider the Reinforcement Learning setting where an agent's interaction with the environment is modeled as a Markov Decision Process (MDP). An MDP is a tuple $(\mathcal{S}, \mathcal{A}, P, r,s_0)$ with state-space $\mathcal{S}$, action-space $\mathcal{A}$, dynamics $P:\mathcal{S} \times \mathcal{A}  \rightarrow\mathcal{S}$, reward function $r(s,a):\mathcal{S} \times \mathcal{A}  \rightarrow\mathbb{R}$ and initial state  $s_0$. $P(.|s,a)$ is the transition probability distribution and $r(s,a)\in [0,R_{\text{max}}]$. We focus on the special case of constrained Markov decision processes (CMDP) \citep{altman1999constrained}, which is an augmented version of MDP with additional costs and trajectory-based constraints. A CMDP is represented as a tuple $(\mathcal{S}, \mathcal{A}, P, r, c, s_0, d_0)$. The terms $\mathcal{S}, \mathcal{A}, P, r, s_0$ are the same as in the unconstrained MDP; the additional terms $c(s)$ is the immediate cost and $d_0\in\mathcal{R}_{\ge0}$ is the maximum allowed value for the expected cumulative cost of the policy. We define a generic version of the Bellman operator w.r.t policy $\pi$ and a function $h(s,a):\mathcal{S} \times \mathcal{A}  \rightarrow\mathbb{R} $ as follows:
\begin{equation}
    \mathcal{B}_{\pi,h}[V][s] = \sum_{a}\pi(a|s)[h(s,a)+\gamma\sum_{s'\in S}P(s'|s,a)V(s')]
\end{equation}

The function $h(s,a)$ can be instantiated to be the reward function or the cost function. When it is the reward function, this becomes the normal Bellman operator in RL. When $h(s,a)$ is replaced by the cost function $c(s)$, it becomes the Bellman operator over the cost objective, which will be used later in designing the Lyapunov function.
We further define $J_{\pi}(s_0)=\E{}{\sum_{t=0}^\infty\gamma^t r(s_t,a_t)|s_0,\pi}$ to be the performance of the policy $\pi$, $D_{\pi}(s_0)=\E{}{\sum_{t=0}^\infty\gamma^t c(s_t,a_t)|s_0,\pi}$ to be the expected cumulative cost of the policy $\pi$, where $\pi$ belongs to the set of stationary policies $\mathcal{P}$. Given a CMDP, we are interested in finding a solution to the following constrained optimization problem:
\begin{equation}
    \label{eq:CMDP}
    \text{max}_{\pi\in\mathcal{P}}[J_{\pi}(s_0)]~\text{s.t}~D_{\pi}(s_0)\le d_0
\end{equation}

% As mentioned in~\citet{chow2018lyapunov}, the objective is formulated with a single start state $s_0$, but the derivations can generalize to $s_0$ that is sampled from a distribution $P_{s_0}$.

% We note that our results can generalize to  distribution of initial states, rewards and costs, but for simplicity we will consider deterministic cost functions and initial state going forward.

\subsection{Safe Reinforcement Learning using Lyapunov Functions}

We will build upon the Lyapunov framework introduced by \cite{chow2018lyapunov}, also known as SDDPG. It proposes to use Lyapunov functions to derive a policy improvement procedure with safety guarantees. The basic idea is that, given a safe baseline policy $\pi_B$, it finds a set of safe policies based on $\pi_B$ using Lyapunov functions. For each policy improvement step, it will then choose the policy with the best performance within this set. 

The method works by first designing a Lyapunov function for a safe update around the current safe baseline policy $\pi_B$. A set of Lyapunov functions is defined as follows:
\begin{equation}
    \label{eq:lyp}
    \mathcal{L}_{\pi_B}(s_0,d_0) = \{L:S\rightarrow R_{\ge0}:\mathcal{B}_{\pi_B,c}[L](s)\le L(s),\forall s \in S; L(s_0)\le d_0 \}
\end{equation}
The Lyapunov functions in this set are designed in a way to construct provably safe policy updates. Given any Lyapunov function within this set $L_{\pi_B}\in \mathcal{L}_{\pi_B}(s_0,d_0)$, we define the set of policies that are consistent with it to be the $L_{\pi_B}$-induced policies:
\begin{equation}
    \mathcal{I}_{L_{\pi_B}} = \{\pi\in\mathcal{P} : \mathcal{B}_{\pi,c}[L_{\pi_B}](s)\le L_{\pi_B}(s), \forall s\in\mathcal{S}\}
\end{equation}
It can be shown that any policy $\pi$ in the $L_{\pi_B}$-induced policy set is ``safe", i.e $D_{\pi}(s_0)\le d_0$~\citep{chow2018lyapunov}.
% \wenxuan{I think this whole part of derivation should go in appendix?}
% It can be derived that any policy $\pi$ in the $L_{\pi_B}$-induced policy set are ``safe": Consider $L_{\pi_B}$ to be an ``initialization" for the cost value function of policy $\pi$. Since the Bellman operator $\mathcal{B}_{\pi,c}$ is a contraction mapping, applying $\mathcal{B}_{\pi,c}$ over any initialization of the cost value function will give us the true value function:
% \begin{equation}
%     D_{\pi}(s) = \lim_{k\rightarrow\infty} \mathcal{B}^{k}_{\pi,c}[L_{\pi_B}](s)~\forall s \in S
% \end{equation}

% According to Eq \ref{eq:lyp}, we have:
% \begin{equation}
%     \mathcal{B}^{k}_{\pi,c}[L_{\pi_B}](s)\le L_{\pi_B}(s)~\forall s \in S, k>0
% \end{equation}

% Thus, $D_{\pi}(s)\le L_{\pi_B}(s)~\forall s \in S $. Since $L_{\pi_B}(s_0)\le d_0$ from Eq. \ref{eq:lyp}, we have $D_\pi(s_0)<d_0$ which is the safety constraint in the CMDP objective (Eq. \ref{eq:CMDP}). In this way, searching the highest performing policy within this $L_{\pi_B}$-induced policy set leads to a safe policy improvement.

% The aim here then is to design a Lyapunov function which contains the optimal policy, i.e optimal policy belongs to the set of L-induced policies so that the optimization restricted in this set indeed results in the solution of Eq. \ref{eq:CMDP}.

The choice of $L_{\pi_B}$ affects the $L_{\pi_B}$-induced policy set. We need to construct $L_{\pi_B}$ such that the $L_{\pi_B}$-induced policy set contains the optimal policy $\pi^*$. % In general, the optimal policy $\pi^*$ does not belong the policies induced by the Lyapunov functions. We define $\pi_B$ to be the safe policy at the current iteration, also called baseline policy.
\cite{chow2018lyapunov} show that one such Lyapunov function is $L_{\pi_B,\epsilon}(s) = \E{}{\sum_{t=0}^{\infty} \gamma^t (c(s_t)+\epsilon(s_t))|\pi_B, s}$, where $\epsilon(s_t)\ge0$. The function $L_{\pi_B,\epsilon}(s)$  can be thought of as a cost-value function for policy $\pi_B$ augmented by an additional per-step cost $\epsilon(s_t)$. Accordingly, we can define the following state-action value function:

\begin{equation}
    \label{eq:lyapunov_q}
    Q_{L_{\pi_B,\epsilon}}(s,a)=c(s)+\epsilon(s)+\gamma \sum_{s'}P(s'|s,a)L_{\pi_B,\epsilon}(s')
\end{equation}

% It can be verified that $\pi_B$ is in the set of $L_{\pi_B,\epsilon}(s)$-induced policies.
% \begin{equation}
%     L_{\pi_B,\epsilon}(s) = \mathcal{B}_{\pi_B,c+\epsilon}[L_{\pi_B,\epsilon}](s) \ge \mathcal{B}_{\pi_B,c}[L_{\pi_B,\epsilon}](s)~~~\text{($\epsilon(s_t)>0~\forall s_t$)}.
% \end{equation}
It was shown in \cite{chow2018lyapunov} that finding a state dependent function $\epsilon$ such that the the optimal policy is inside the corresponding $L_{\pi_B,\epsilon}$-induced set is generally not possible and requires knowing the optimal policy. As an approximation, they suggest to create the Lyapunov function with the largest auxiliary cost $\hat{\epsilon}$, such that $ L_{\pi_B,\hat{\epsilon}}(s) \ge \mathcal{B}_{\pi_B,c}[L_{\pi_B,\hat{\epsilon}}](s) $ and $L_{\pi_B,\hat{\epsilon}}(s_0)\le d_0$. A larger auxiliary cost $\epsilon$ per state ensures that we have a larger set of L-induced policies, making it more likely to include the optimal policy in the set. The authors show that the following $\hat{\epsilon}(s)$ in the form of a constant function satisfies the conditions described:
\begin{equation}
    \label{eq:constraint_budget}
    \hat{\epsilon}(s)= (1-\gamma)(d_0-D_{\pi_B}(s_0))
\end{equation}

Plugging this function $\hat{\epsilon}(s)$ and the definition of $Q_{L_{\pi_B,\epsilon}}(s,a)$ into the CMDP objective, the policy update under the set of policies that lie in the $L_{\pi_B,\hat{\epsilon}}$-induced policy set, or equivalently the policies that are safe, is given by:
\begin{equation}
    \label{lyapunov_constraint}
    \pi_+(.|s) =  \max_{\pi\in\mathcal{P}} J_{\pi}(s_0),~s.t~\int_{a\in \mathcal{A}}(\pi(a|s)-\pi_B(a|s)) Q_{L_{\pi_B,\hat{\epsilon}}}(s,a) da\le \hat{\epsilon}(s)~\forall s \in \mathcal{S}
\end{equation}

In the case of a deterministic policy, the policy update becomes:

\begin{equation}
\label{eq:deterministic_lyapunov}
    \pi_+(.|s) =  \max_{\pi\in\mathcal{P}} J_{\pi}(s_0),~\text{s.t}~ Q_{L_{\pi_B,\hat{\epsilon}}}(s,\pi(s))-Q_{L_{\pi_B,\hat{\epsilon}}}(s,\pi_B(s)) \le \hat{\epsilon}(s)~\forall s \in \mathcal{S}
\end{equation}

We build upon this objective in our work. We include the proof of the Lyapunov approach for completeness in Appendix~\ref{ap:lyapunov_derivation}, and we advise the reader to see previous work \citep{chow2018lyapunov} for a more detailed derivation. Using the Lyapunov function, the trajectory-based constraint of the CMDP is converted to a per-state constraint (Eq.~\ref{lyapunov_constraint}), which is often much easier to deal with.

\section{Method}

\subsection{Barrier function for Lyapunov Constraint}

We present Lyapunov Barrier Policy Optimization (LBPO) that aims to update policies inside the $L_{\pi_B,\hat{\epsilon}}$-induced policy set. We work under the standard policy iteration framework which contains two steps: Q-value Evaluation and Safe Policy Improvement. We initialize LBPO with a safe baseline policy $\pi_B$. In practice, we can obtain safe initial policies using a simple (usually poorly performing) hand-designed control policy; in our experiments, we simplify this process and achieve safe initial policies by training on the safety objective. We assume that we have $m$ different constraints, as LBPO naturally generalizes to more than one constraint. 

\subsubsection{Q Evaluation}
We use on-policy samples to evaluate the current policy. We compute a reward Q function $Q^R$, and cost Q functions $Q^{C_i}$ corresponding to each cost constraint $i \in [1,2...m]$. Each Q function is updated using TD($\lambda$)~\citep{sutton1988learning} which helps us more accurately estimate the Q functions.  Furthermore, we use the on-policy samples to get the cumulative discounted cost $D^i_{\pi}(s_0)$ of the current policy, which allows us to set up the constraint budget for each constraint given by $\epsilon_i = (1-\gamma) (d_0^i-D^i_{\pi}(s_0))$ as shown in Eq. \ref{eq:constraint_budget}.

\subsubsection{Regularized Safe Policy Update}
In this work, we focus on deterministic policies, where we have the following policy update under the $L$-induced set for each constraint as given in Eq.~\ref{eq:deterministic_lyapunov} :\\
\begin{equation}
    \pi_+(.|s) =  \max_{\pi\in\mathcal{P}} J_{\pi}(s_0)~\text{s.t}~ Q^i_{L_{\pi_B,\hat{\epsilon}}}(s,\pi(s))-Q^i_{L_{\pi_B,\hat{\epsilon}}}(s,\pi_B(s)) \le \hat{\epsilon}_i(s)~\forall i\in[1,2,...m],\forall s \in \mathcal{S}
\end{equation}

% Appendix
% Using the definition of $Q^i_L_{\pi}(s,a)$ from Eq.~\ref{eq:lyapunov_q}  and when $\hat{\epsilon}(s)$ is a constant function (denote by $\hat{epsilon}$), we can replace $Q^i_{L_{\pi_B}}$ by $Q^{C_i}$,
% \begin{align*}
%   Q_{L_{\pi_B}}(s,a)&=c(s)+\hat{\epsilon}+\gamma \sum_{s'}P(s'|s,a)L^{\pi_B}_{\hat{\epsilon}}(s')\\
%   &= c(s)+\hat{\epsilon}+\gamma \sum_{s'}P(s'|s,a)[c(s')+\hat{\epsilon}+\sum_{s''}P^{\pi_B}_{(s''|s')}(L^{\pi_B}_{\hat{\epsilon}}(s''))]\\
%   &= \sum_{t=0}^{\infty}\gamma^t\hat{\epsilon} +  \E{}{\sum_{t=0}^{\infty}\gamma^t c(s_t)|\pi_B,a_0=a, s_0= s}\\
% &= \sum_{t=0}^{\infty}\gamma^t\hat{\epsilon} +  Q^C_{\pi}(s,a)
% \end{align*}
% which is the cost Q function for the $i^{th}$ cost constraint, since the Lyapunov function $Q^i_L_{\pi}(s,a)$ and the cost-Q function $Q^{C_i}_{\pi}(s,a)$ only differ by a constant ($\sum_{t=0}^\infty \gamma^t\hat{\epsilon}$).

% since the Lyapunov function $Q^i_L_{\pi}(s,a)$ and the cost-Q function $Q^{C_i}_{\pi}(s,a)$ only differs by a constant ($\sum_{i=0}^\infty \gamma^i\hat{\epsilon}$), we can replace $Q^i_{L_{\pi_B}}$ by $Q^{C_i}$, which is the cost Q function for $i^{th}$ cost.
We can simplify this equation further by replacing $Q^i_{L_{\pi_B,\hat{\epsilon}}}$ with $Q^{C_i}_{\pi_B}$ which is the $i^{th}$ cost Q-function under the policy $\pi_B$, when $\epsilon$ is a constant function (see Appendix \ref{ap:lyapunov_to_costq}).  To ensure that the Lyapunov constraints are satisfied, we construct an unconstrained objective using an indicator penalty $I(Q^{C_{i}}_{\pi_B}(s,\pi_{\theta}(s)))$ for each constraint. 
\begin{equation}
    \label{eq:indicator_barrier}
    I(Q^{C_{i}}_{\pi_B}(s,\pi_{\theta}(s))) = \begin{cases}
    0 &\text{$Q^{C_{i}}_{\pi_B}(s,\pi_{\theta}(s))-Q^{C_{i}}_{\pi_B}(s,\pi_B(s)) \le \hat{\epsilon}_i(s)$}\\
    \infty &\text{$Q^{C_{i}}_{\pi_B}(s,\pi_{\theta}(s))-Q^{C_{i}}_{\pi_B}(s,\pi_B(s)) > \hat{\epsilon}_i(s)$}
\end{cases}
\end{equation}
% In case the constraint is satisfied, we optimize for the unconstrained performance objective; otherwise, the policy must update such that it stays within the constraints as it gets a penalty of $\infty$ outside the feasible region. 
We will use a differentiable version of the indicator penalty called the logarithmic barrier function: 
\begin{equation}
\label{eq:logarithmic_barrier}
        \psi(Q^{C_{i}}_{\pi_B}(s,\boldsymbol{\pi_{\theta}(s))}) = - 
        \beta\log\bigg(  \hat{\epsilon}(s) - \big(Q^{C_{i}}_{\pi_B}(s,\boldsymbol{\pi_{\theta}(s)})-Q^{C_{i}}_{\pi_B}(s,\pi_B(s)\big) \bigg)
\end{equation}
%         \beta{\log\bigg(Q^{C_{i}}_{\pi_B}(s,\pi_{\theta}(s))-Q^{C_{i}}_{\pi_B}(s,\pi_B(s)) \- \hat{\epsilon}(s)\bigg)}
The function $\psi$ is parameterized by $\theta$ and $Q^{C_{i}}_{\pi_B}(s,\pi_B(s))$ is a constant. Our policy update will use the gradient at $\pi_\theta=\pi_B$, ensuring that the logarithmic barrier function is well defined, since $\hat{\epsilon}(s)>0~\forall s$.

% \hs{Refine: Since we evaluate $Q^{C_{i}}_{\pi_B}(s,\boldsymbol{\pi_{\theta}(s)})$ at $\pi_\theta=\pi_B$, the second term $Q^{C_{i}}_{\pi_B}(s,\boldsymbol{\pi_{\theta}(s)})$ and the third term $Q^{C_{i}}_{\pi_B}(s,\pi_B(s))$ have the same value. However, the gradient only pass through the second term while $Q^{C_{i}}_{\pi_B}(s,\pi_B(s))$ is treated as a constant.} % $Q^{C_{i}}_{\pi_B}(s,\pi_B(s))$ cancels out the value of the $Q^{C_{i}}_{\pi_B}(s,\pi_{\theta}(s))$ such that $\hat{\epsilon}$ limits the change of the Q-function and is invariant to its value.

% Thus, we can use first-order gradient methods with a sufficiently small trust region to minimize the error in verifying the constraint violation. 

\begin{wrapfigure}{R}{0.4\textwidth}
    \includegraphics[width=0.9\linewidth]{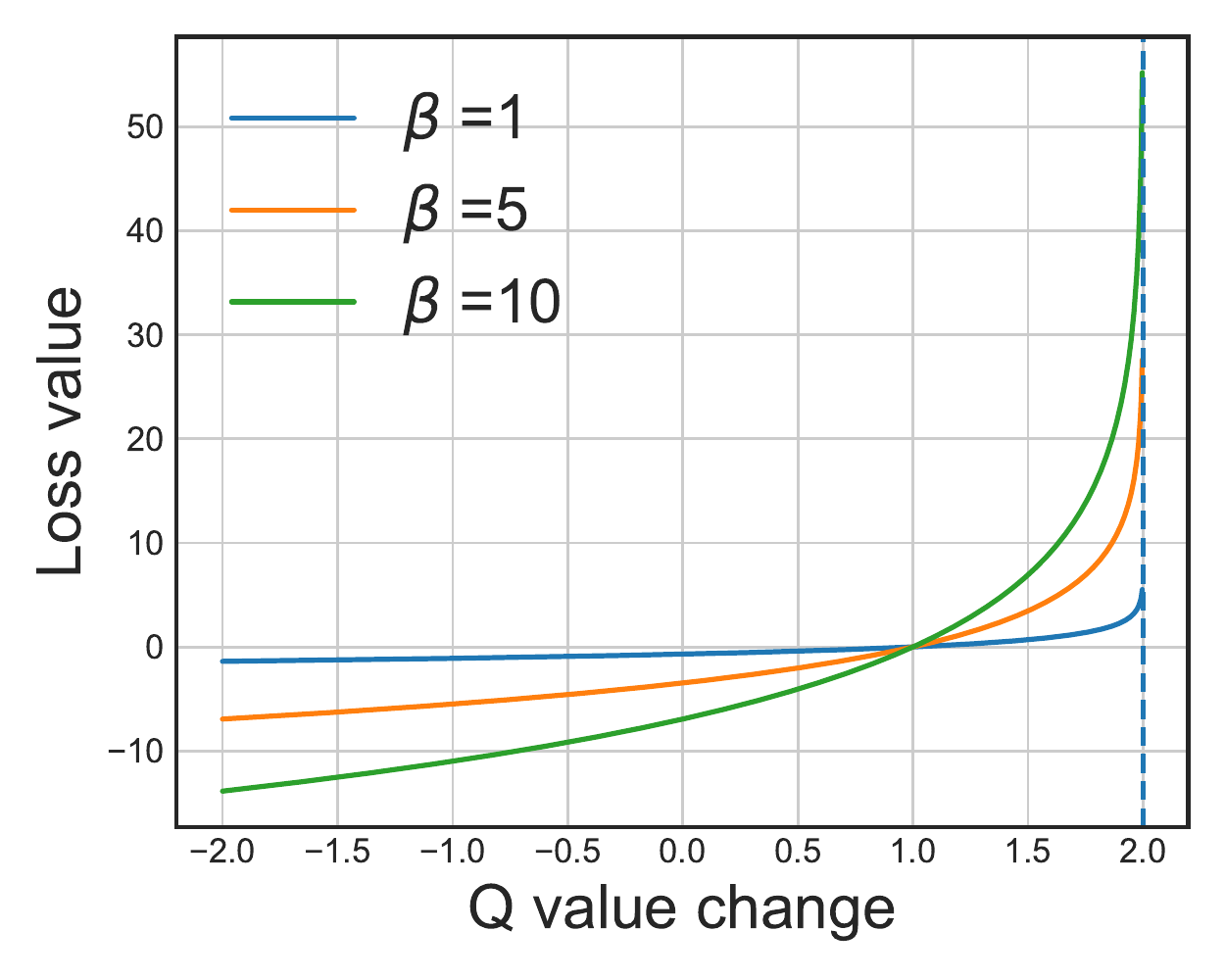}
 \caption{As the difference between Q values for action a and the baseline action reaches $\epsilon$ (in this case $\epsilon=2$), the loss increases to $\infty$.}
    \label{fig:barriers}
    \vspace{-1mm}
\end{wrapfigure}

Figure~\ref{fig:barriers} captures the behavior of the function $\psi$ for different $\beta$. The parameter $\beta$ captures the amount of risk-aversion we desire from the agent. A high $\beta$ will help avoid constraint violations occurring due to approximation errors in our learned Q-functions. We verify this empirically in Section~\ref{sec:experiments}.

% In this way, we can incorporate the constraint into the optimization objective for the policy update by adding $\psi(Q^{C_{i}}_{\pi_B}(s,\pi_{\theta}(s)))$. Note that in the barrier function, 

% \begin{align}
% \label{eq:policy_objective}
%     \pi =& \text{max}_{\pi\in\mathcal{P}} J_{\pi}(s_0) - \E{s\sim\mathcal{R}}{\sum_{i=1}^m \psi(Q^{C_i}(s,\pi))}
% \end{align}

We use the Deterministic Policy Gradient Theorem~\citep{silver2014deterministic} for the policy update. For updating a $\theta$-parameterized policy with respect to the expected return, the objective can be written as: $\text{argmin}_{\theta}~\E{s \sim \mathcal{\rho^{\pi_B}}}{(-Q^R_{\pi_B}(s,\pi_{\theta}(s)))}$, where $\rho^{\pi_B}$ is the on-policy state distribution. 

Since we rely on on-policy samples for Q-function evaluation, the Q-function estimation outside the on-policy state distribution can be arbitrarily bad. Similar to~\cite{schulman2015trust}, we constrain our policy update using a hard KL constraint (i.e. a trust region) between the current policy and the updated policy under the presence of stochastic exploration noise. The trust region also allows us to make sure that our policy change is bounded, which allows us to ensure that with a small enough trust region, our first order approximation of the objective is valid.

%gradient update is accurate enough to reason about the cases in Eq.~\ref{eq:indicator_barrier}.

Augmenting the  on-policy update with the Lyapunov barrier and a KL regularization, we have the following LBPO policy update:
\begin{align}
\label{eq:LBPO_objective}
    \theta_{k+1} =&~\text{argmin}_{\theta}~ \E{s\sim\rho^{\pi_B}}{-Q^R_{\pi_B}(s,\pi_\theta(s)) + \sum_{i=1}^m \psi(Q^{C_i}_{\pi_B}(s,\pi_\theta(s)))}\\
    &\text{subject to}~\E{s\sim\rho^{\pi_B}}{D_{\text{KL}}(\pi_\theta[.|s]+\mathcal{N}(0,\delta)~\|~\pi_B[.|s]+\mathcal{N}(0,\delta))}< \mu
\end{align}
where $\delta$ is the exploration noise, $\rho^{\pi_B}$ is the state distribution induced by the current policy, $\mu$ is the expected KL constraint threshold and we set $\pi_B$ to the safe policy at iteration $k$, as the update guarantees safe policies at each iteration. In practice, we expand our objective using a Taylor series expansion and solve to a leading order approximation around $\theta_k$. Letting the gradient of the objective in Eq~\ref{eq:LBPO_objective} be denoted by $g$ and the Hessian of the KL divergence by $H$, our objective becomes:
\begin{align}
    \theta_{k+1} &= \text{argmin}_\theta~ g^T(\theta-\theta_k),
    ~~~ \text{subject to}~\frac{1}{2}(\theta-\theta_k)^T H(\theta-\theta_k)\le\mu
    \vspace{-2mm}
\end{align}
We solve this constrained optimization using the Fisher vector product with Conjugate gradient method similar to~\cite{schulman2015trust}.
\section{Experiments}
\label{sec:experiments}
In this section, we evaluate LBPO and compare it to prior work. First, we benchmark our method against previous baselines to show that LBPO can achieve better constraint satisfaction while being competitive in performance. Second, we give empirical evidence that previous methods near constraint satisfaction can be explained by backtracking. Third, we show by a didactic example that LBPO is more robust than CPO and SDDPG to Q-function errors, hence making it a preferable alternative, especially when function approximation is used. Finally, we show that LBPO allows flexible tuning of the amount of risk-aversion for the agent. 

% \begin{enumerate}
% \item How does LBPO compare to other methods for solving CMDP's?
% \item  Does LBPO allows the agent to be more conservative depending on the barrier? 
% \item does LBPO compare to baselines like CPO, SDDPG in avoiding constraint violations rather than correcting in? 
% \end{enumerate}
\begin{figure}\centering
    \includegraphics[width=1.0\textwidth]{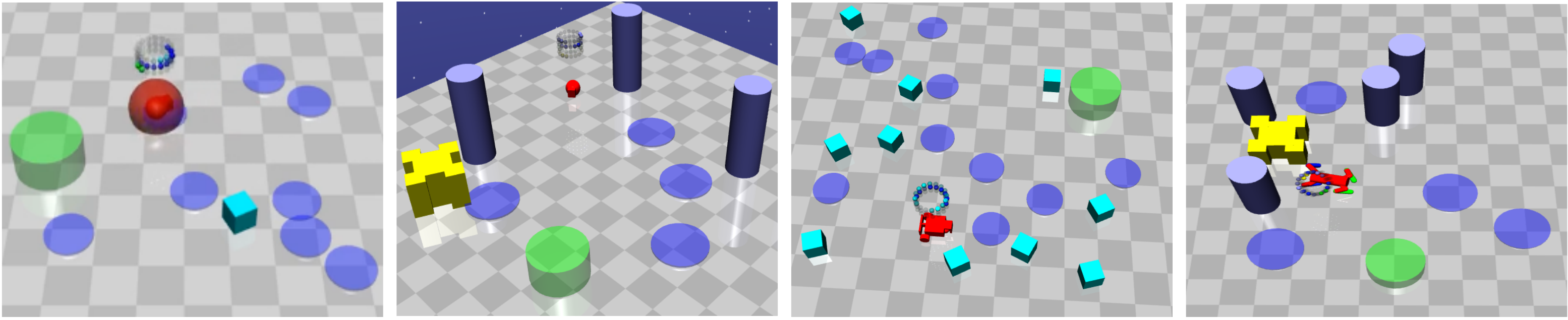}
 \caption{OpenAI Safety Environments: PointGoal1, PointPush2, CarGoal2, DoggoPush2}
    \label{fig:safety-environments}
    \vspace{-5mm}
\end{figure}

\begin{wrapfigure}{R}{0.5\textwidth}
    \centering
    \includegraphics[width=1.0\linewidth]{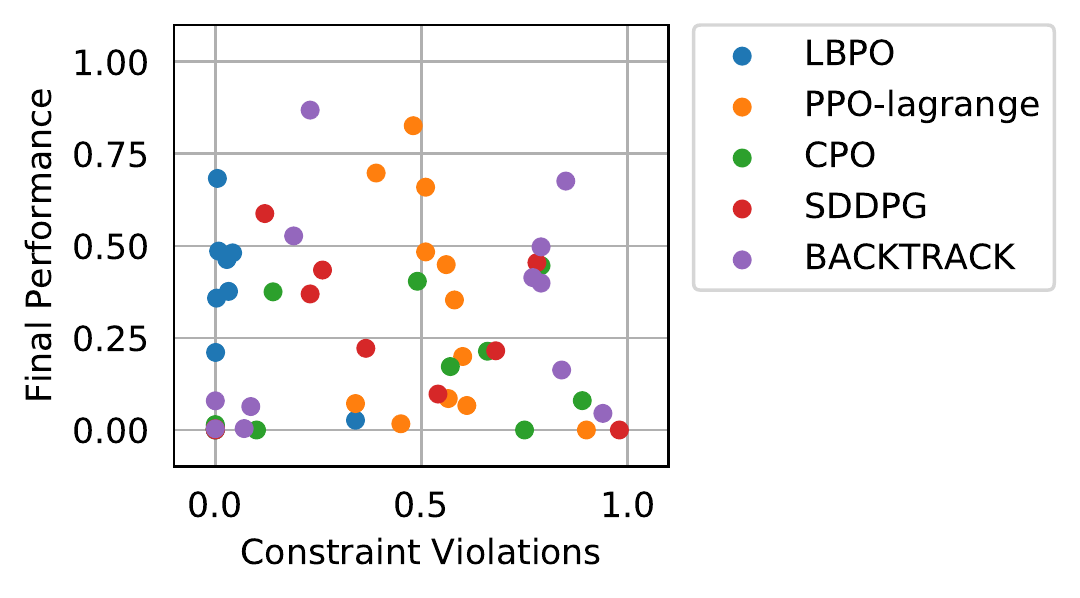}
 \caption{Each point corresponds to a particular safety method applied to a certain safety environment. The x-axis shows the fraction of constraint violations encountered by the behavior policy during training and y-axis shows the policy performance normalized by the corresponding environment's PPO return.}
 \vspace{2mm}
 \label{fig:cost_vs_reward}
\end{wrapfigure}

\textbf{Comparisons.} For our experiments, we compare LBPO against a variety of baselines: PPO, PPO-lagrangian, SDDPG, CPO and BACKTRACK. PPO~\citep{schulman2017proximal} is an on-policy RL algorithm that updates in an approximate trust-region without considering any constraints. PPO-lagrangian belongs to a class of Lagrangian methods~\citep{altman1998constrained} for safe Reinforcement Learning, which transforms the constrained optimization problem to a penalty form $ \text{max}_{\pi\in \mathcal{P}}\text{min}_{\lambda\ge0}\mathbb{E}[{\sum_{t=0}^\infty \gamma^tr(s_t,a_t)} + $ $\lambda(\sum_{t=0}^\infty \gamma^t c(s_t)-d_0)|\pi,s_0]$.
% \begin{equation}
% \footnotesize
%     \text{max}_{\lambda\ge0}\text{min}_{\pi\in \mathcal{P}}\E{}{\sum_{t=0}^\infty \gamma^t(r(s_t,a_t)+ \lambda c(s_t))|\pi,s_0}
% \end{equation}
$\pi$ and $\lambda$ are jointly optimized to find a saddle point of the penalized objective. SDDPG~\citep{chow2018lyapunov,chow2019lyapunov} introduces the Lyapunov framework for safe-RL and proposes an action-projection  method which in theory guarantees the update of the policy within a safe set. We evaluate the $\alpha$-projection version of SDDPG~\citep{chow2019lyapunov}. Since the original implementation for the method is unavailable, we re-implemented the method to the best of our abilities. CPO~\citep{achiam2017constrained} derives a trust-region update rule which guarantees the monotonic improvement of the policy while satisfying constraints. CPO also uses a backtracking recovery rule. We elaborate on the BACKTRACK baseline in Section~\ref{exp:backtracking_baselines}.

\textbf{Tasks}. We evaluate these methods using the OpenAI Safety Gym~\citep{Ray2019}, which consists of 12 continuous control MuJoCo tasks~\citep{todorov2012mujoco}. These tasks use 3 robots: Point, Car, and Doggo. Point is the simplest of three which can be commanded to move forward/backward or to turn. Car has two driven wheels which needs to be controlled together to obtain forward/backward and turning behavior. Doggo is a quadrupedal robot whose joint angles at hip, knee and torso can be commanded to obtain similar behavior. Each robot has 2 types of tasks  (Goal, Push) with 2 difficulty levels  (1, 2). In Goal tasks, the robot has to move to a series of goal locations, and in Push tasks, the robot has to push a box to a series of goal locations. There are mobile and immobile obstacles made up of a hazard region, vases and pillars which generate a cumulative penalty for the agent. Point has an observation space of 60 dimensions, Car has 72 dimensions, and Doggo has 104 dimensions, which constitute sensor readings, joint angles, and velocities. The environments are shown in Figure~\ref{fig:safety-environments}. 

%We also show in the Appendix that CPO relies heavily on  backtracking during training.

% We demonstrate the validity of our hypothesis by observing that BACKTRACK has competitive performance with CPO and by showing that CPO relies on a large number of backtracking behavior during its training process.

% \begin{figure}[t]
%     \centering
%     \includegraphics[width=\linewidth]{figures/full_result.pdf}
%     \caption{Training curved for LBPO in comparison to baselines: PPO, PPO-lagrangian, CPO, SDDPG. We also compare against our simple baseline BACKTRACK here. For each environment, the top row shows the Average undiscounted cumulative cost during training, and bottom row shows the Average undiscounted return. PPO often has large constraint violations and is clipped from some plots, when its constraint violations are high. Red dashed line in Average Cost plots shows the constraint limit which is 25 in all environments.}
%     \label{fig:lbpo_benchmark}
% \end{figure}

\subsection{Safe Reinforcement Learning Benchmarks}

We summarize the comparison of LBPO to all of the baselines (PPO, PPO-lagrangian, BACKTRACK, SDDPG and CPO) on the OpenAI safety benchmarks in Figure~\ref{fig:cost_vs_reward} and Tables ~\ref{tab:mujoco_all_constraint} and~\ref{tab:mujoco_all_reward}. Additional training plots for policy return and policy cost can be found in Appendix~\ref{ap:benchmarks}.

\begin{table}[h]
    \vspace{-2mm}
    \centering
    \footnotesize
    % \resizebox{\textwidth}{!}{%
    \begin{tabular}{c|c|c|c|c|c|c}
        \toprule
         Method & PPO & PPO-lagrangian & CPO & SDDPG& BACKTRACK & LBPO\\
        \midrule
        PointGoal1 & 1.00 & 0.48 & 0.79 &0.78 & 0.85 & \textbf{0.04} \\
        PointGoal2 & 0.98 & 0.60 & 0.75 &0.98 & 0.94 & \textbf{0.34} \\
        PointPush1 & 1.00 & 0.51 & 0.14 & 0.12 & 0.19 & \textbf{0.00} \\
        PointPush2 & 1.00 & 0.51 & 0.66 & 0.36 & 0.77 & \textbf{0.00} \\
        CarGoal1 & 1.00 & 0.56 & 0.89& 0.54 & 0.79 & \textbf{0.03} \\
        CarGoal2 & 1.00 & 0.61 & 0.57 & 0.68 & 0.84 & \textbf{0.00} \\
        CarPush1&0.99 & 0.39 & 0.10 & 0.26 & 0.23 & \textbf{0.01} \\
        CarPush2 & 1.00 & 0.58 & 0.49 & 0.23 & 0.79 & \textbf{0.03} \\
        DoggoGoal1 & 1.00 & 0.90 & \textbf{0.00} & \textbf{0.00} & 0.07 & \textbf{0.00} \\
        DoggoGoal2 & 1.00 & 0.45 & \textbf{0.00} & \textbf{0.00} & \textbf{0.00} & \textbf{0.00} \\
        DoggoPush1 & 0.98 & 0.56 & \textbf{0.00} & \textbf{0.00} & \textbf{0.00} & \textbf{0.00} \\
        DoggoPush2& 1.00 & 0.34 & \textbf{0.00} & \textbf{0.00} & 0.09 & \textbf{0.00} \\
        \bottomrule
    \end{tabular}
    % }
    \caption{We report the fraction of unsafe behavior policies encountered during training across different OpenAI safety environments for the policy updates across $2e^7$ training timesteps. LBPO obtains fewer constraint violations consistently across all environments.}
    \label{tab:mujoco_all_constraint}
    \vspace{2mm}
\end{table} 
\begin{table}[h]
    \vspace{-2mm}
    \centering
    \footnotesize
    % \resizebox{\textwidth}{!}{%
    \begin{tabular}{c|c|c|c|c|c|c}
        \toprule
         Method & PPO & PPO-lagrangian & CPO & SDDPG& BACKTRACK & LBPO\\
        \midrule
        PointGoal1 &1.00&\textbf{0.826}&0.450&0.451&0.670&0.480\\
        PointGoal2 &1.00&\textbf{0.200}&0.000&0.000&0.045&0.026\\
        PointPush1 & 1.00 & 0.659 & 0.375 & 0.587 & 0.527 & \textbf{0.683} \\
        PointPush2 & 1.00 & \textbf{0.483} & 0.213 & 0.221 &0.413 & 0.358 \\
        CarGoal1 & 1.00 & 0.449 & 0.079 & 0.097 & \textbf{0.497} & 0.376 \\
        CarGoal2 & 1.00 & 0.066 & 0.172 & \textbf{0.215} & 0.162 & 0.210 \\
        CarPush1 & 1.00 & 0.697 & 0.000 & 0.434 & \textbf{0.868} & 0.485 \\
        CarPush2 & 1.00 & 0.353 & 0.403 & 0.369 & 0.399 & \textbf{0.430} \\
        DoggoGoal1 &1.00 & 0.000 & 0.003 & 0.002 & 0.003 & \textbf{0.007} \\
        DoggoGoal2 &1.00 & \textbf{0.016} & 0.002 & 0.003 & 0.003 & 0.003 \\
        DoggoPush1 & 1.00 & \textbf{0.080} & 0.014 & 0.001 & 0.079 & 0.012 \\
        DoggoPush2 & 1.00 & \textbf{0.071} & 0.000 & 0.000 & 0.063 & 0.000 \\
        \bottomrule
    \end{tabular}
    % }
    \caption{Cumulative return of the converged policy for each safety algorithm normalized by PPO's return. Negative returns are clipped to zero. LBPO tradeoffs return for better constraint satisfaction. Bold numbers show the best performance obtained by a safety algorithm (thus excluding PPO).}
    \label{tab:mujoco_all_reward}
    \vspace{-6mm}
\end{table} 

\textbf{Constraint Satisfaction}. Table~\ref{tab:mujoco_all_constraint} shows that in all the environments, LBPO actively avoids constraint violations, staying below the threshold in most cases. In the PointGoal2 environment, no method can achieve good constraint satisfaction which we attribute to the nature of the environment as it was found that safe policies were not obtained even when training only on the cost objective. In all the other cases we note that LBPO achieves near-zero constraint violations.

Like our method, SDDPG also builds upon the optimization problem from Equation~\ref{eq:deterministic_lyapunov} but solves this optimization using a projection onto a safe set instead of using a barrier function.  We noticed the following practical issues with this approach: First, in SDDPG, each safe policy is composed of a projection layer, which itself relies on previous safe policies. This requires us to maintain all of the previous policies and thus the memory requirement grows linearly with the number of iterations. SDDG circumvents this issue by using a policy distillation scheme~\citep{chow2018lyapunov}, which behavior clones the safe policy into a parameterized policy not requiring a projection layer. However, behavior cloning introduces errors in the policy leading to frequent constraint violations.  Second, we will show in section~\ref{sec:Robustness to finite sample sizes}  that SDDPG is more sensitive to Q-function errors. PPO-lagrangian produces policies that are only safe \textit{asymptotically} and makes no guarantee of the safety of the behavior policy during each training iteration. In practice, we observe that it often violates constraints during training. 

\textbf{Behavior Policy Performance}. OpenAI safety gym environment provide a natural tradeoff between reward and constraint. A better constraint satisfaction often necessitates a lower performance. We observe in Table~\ref{tab:mujoco_all_reward} that LBPO achieves performance competitive to the baselines. 

%As shown in Section ~\ref{exp:backtracking_baselines}, CPO, and SDDPG inherently rely on backtracking recovery to achieve near constraint satisfaction. 

%PPO-lagrangian was observed to have unstable training performance. 

% \textbf{Backtracking Baseline vs. Other baselines}.
\subsection{Backtracking Baseline}
\label{exp:backtracking_baselines}
CPO~\citep{achiam2017constrained} and SDDPG~\citep{chow2019lyapunov} both use a recovery rule once the policy becomes unsafe, which is to train on the safety objective to minimize the cumulative cost until the policy becomes safe again. In this section, we test the hypothesis that CPO and SDDPG are unable to actively avoid constraint violation but their near constraint satisfaction behavior can be explained by the recovery rule. To this end, we introduce a simple baseline, BACKTRACK, which uses the following objective for policy optimization under a trust region (we use the same trust region as in LBPO):
\begin{equation}
    \pi = \begin{cases}
    \text{max}_{\pi \in \mathcal{P}}~ \E{s\sim\rho^{\pi_B}}{Q^R_{\pi_B}(s,\pi(s))} &\text{if $\pi_B$ is safe}\\
   \text{min}_{\pi \in \mathcal{P}}~ \E{s\sim\rho^{\pi_B}}{Q^C_{\pi_B}(s,\pi(s))} &\text{if $\pi_B$ is unsafe}
\end{cases}\\
\end{equation}
Thus, if the most recent policy $\pi_B$ is evaluated to be safe, BACKTRACK exclusively optimizes the reward; however, if the most recent policy $\pi_B$ is evaluated to be unsafe, BACKTRACK exclusively optimizes the safety constraint.  Effectively, BACKTRACK relies only on the recovery behavior that is used in CPO and SDDPG, without incorporating their mechanisms for constrained policy updates. In Tables~\ref{tab:mujoco_all_constraint} and~\ref{tab:mujoco_all_reward}, we see that BACKTRACK is competitive to both CPO and SDDPG in terms of both constraint satisfaction and performance (maximizing reward), suggesting that the recovery behavior is itself sufficient to explain their performance. In Appendix~\ref{ap:backtracks}, we compare the number of backtracks performed by CPO, SDDPG and BACKTRACK.

\subsection{Robustness to finite sample sizes}
\label{sec:Robustness to finite sample sizes}
% In the presence of finite sample errors in the cost Q function, we observe that CPO is more susceptible to constraint violations compared to LBPO. 
\begin{wrapfigure}{R}{0.5\textwidth}
    \includegraphics[width=0.9\linewidth]{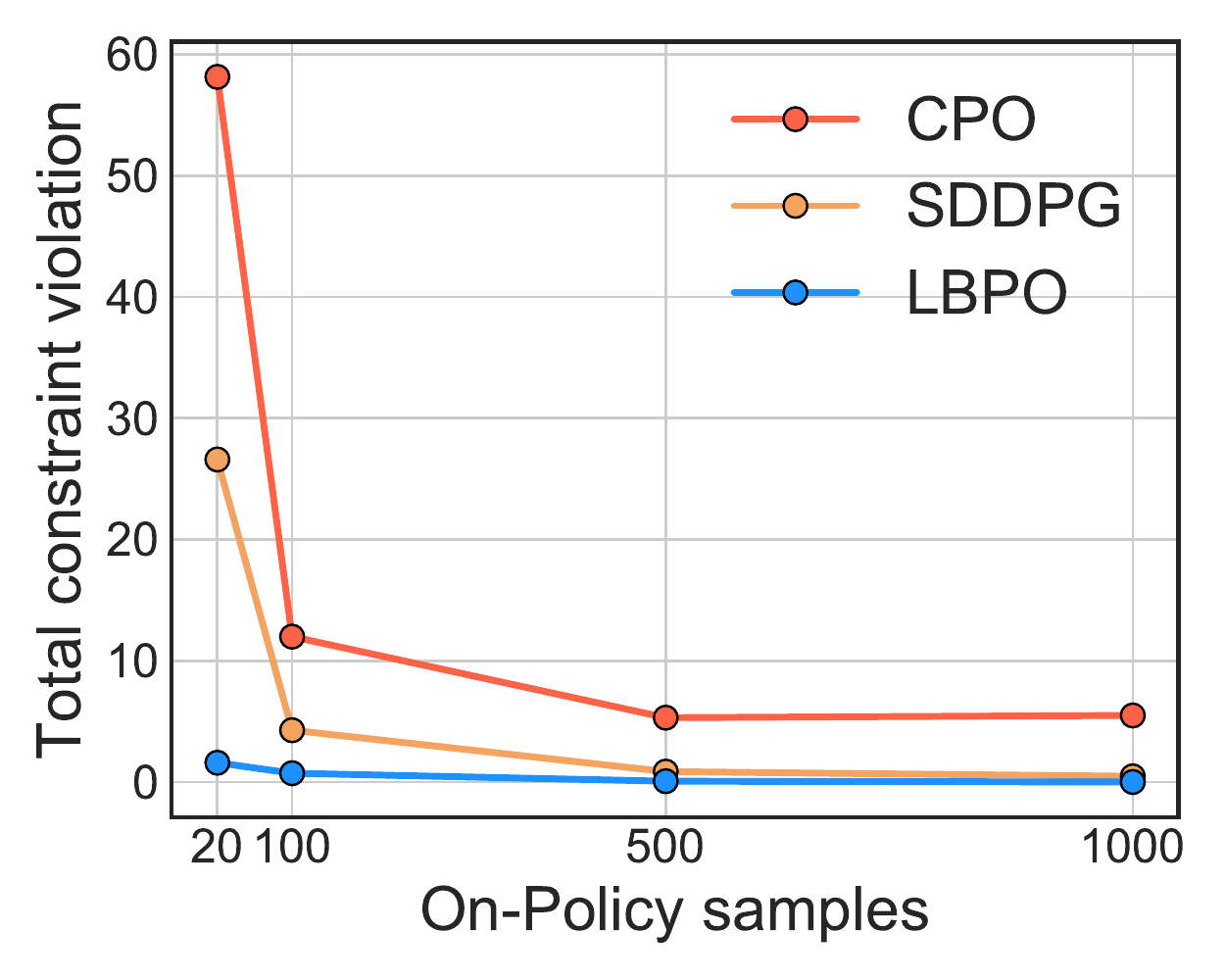}
 \caption{An analysis of the robustness of safe RL algorithms CPO, SDDPG, and LBPO to finite sample Q function errors for a simple didactic environment. Constraint violations in CPO and SDDPG increase quickly as the number of on-policy samples used to estimate the Q function decreases. Results are averaged over 5 seeds.}
 \label{fig:q_analysis}
\end{wrapfigure}
We generally work in the function approximation setting to accommodate high dimensional observations and actions, and this makes it necessary to rely on safety methods that are robust to Q-function errors.  To analyze how robust different methods are to such errors, we define a simple reinforcement learning problem: Consider an MDP with two dimensional state space given by ($x,y$) $\in\mathbb{R}$. The initial state is (0,0). Actions are two dimensional, given by $(a_1,a_2)$ : $a_1,a_2\in[-0.2,0.2]$. The horizon is 10 and the transition probability distribution is $(x',y')=(x,y)+(a_1,a_2)+\mathcal{N}(0,0.1)$. The reward function is $r(x,y)=\sqrt{x^2+y^2}$. The cost function is equal to the reward function for all states, and the constraint threshold is set to 2. We plot in Figure \ref{fig:q_analysis} the total constraint violations during 100 epochs of training with varying number of samples used to estimate the cost Q-function. We find that LBPO is more robust to Q-function errors due to limited data compared to CPO and SDDPG. In this experiment we use $\beta=0.005$, similar to the value used for the benchmark experiments.

\subsection{Tuning conservativeness with the barrier}
\vspace{-2mm}
\begin{figure}\centering
    \includegraphics[width=1.0\textwidth]{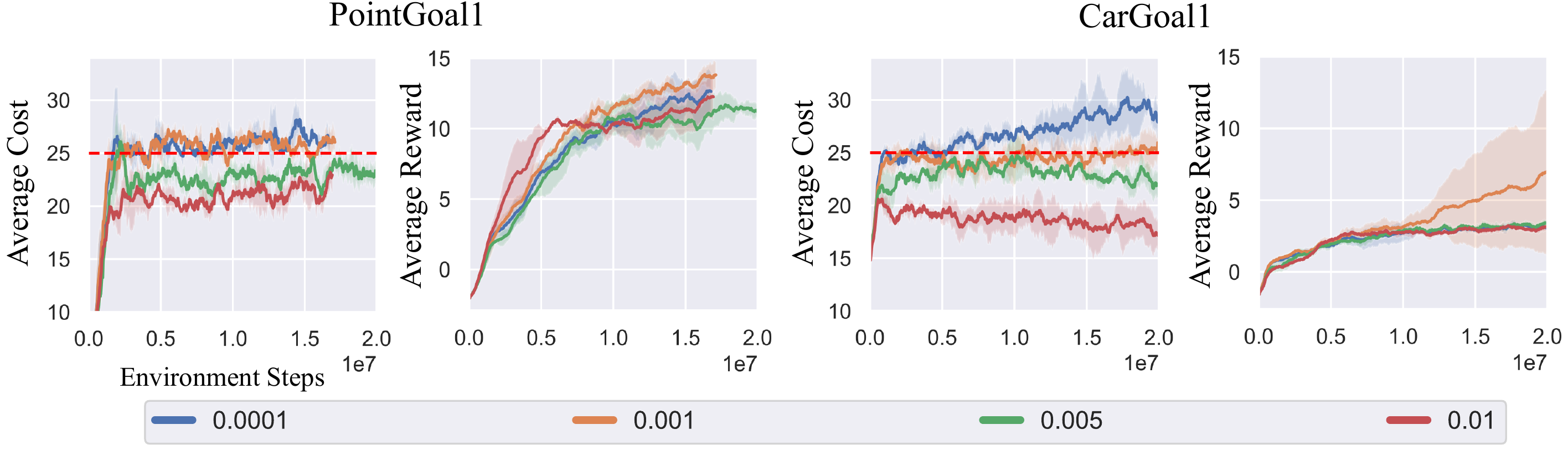}
 \caption{Increasing $\beta$ parameter for the barrier increases the risk aversion of the agent as can be seen in the plots above.  }
    \label{fig:sensitivity}
    \vspace{-5mm}
\end{figure}
A strength of LBPO is the ability to tune  the barrier to adjust the amount of risk-aversion of the agent. Specifically, $\beta$ in Equation~\ref{eq:logarithmic_barrier} can be tuned; a larger $\beta$ leads to more conservative policies. 
% This ability of LBPO may be exploited in cases where we want to learn close to constraint but safe policies, by starting with a large $\beta$ and adapting it according to the training history. 
In Figure \ref{fig:sensitivity}, we empirically demonstrate the sensitivity of $\beta$  to the conservativeness of the policy update. For our benchmark results, we do a hyperparameter search for $\beta$ in the set (0.005, 0.008, 0.01, 0.02) and found that 0.005 works well across most environments.
\section{Related Work}

% \textbf{Model-free RL} Model-free reinforcement learning algorithms achieve high performance for many tasks, but these methods have no way to efficiently deal with real world constraints. Robots in the real world have to be heavily instrumented to ensure that they do not damage themselves and the environment when they train~\citep{dulac2019challenges}. PPO~\citep{schulman2017proximal} and TRPO~\citep{schulman2015trust} are some of the state-of-the-art on-policy reinforcement learning methods which require new samples to be collected for every update to the policy. In this work, we rely on on-policy RL methods for getting accurate cost estimates for each policy encountered during training and use the Deterministic Policy Gradient~\citep{silver2014deterministic} under a trust region for policy updates.   

\textbf{Constrained Markov Decision Process} CMDP's \citep{altman1998constrained} have been a popular framework for incorporating safety in the form of constraints. In CMDP's the agent tries to maximize expected returns by satisfying constraints on expectation of costs. \cite{altman1999constrained} demontrated that for finite MDP with known models, CMDP's can be solved by solving the dual LP program. For large state dimensions (or continuous), solving the LP becomes intractable. A common way to solve CMDP in large spaces is to use the Lagrangian Method~\citep{altman1999constrained,geibel2005risk,chow2017risk}. These methods augment the original RL objective with a penalty on constraint violation and computes the saddle point of the constrained policy optimization via primal-dual methods. These methods give no guarantees of safety during training and are only guaranteed asymptotically at convergence. CPO~\citep{achiam2017constrained} is another method for solving CMDP's that derives an update rule in the trust region which guarantees monotonic policy improvement under constraint satisfaction, similar to TRPO~\citep{schulman2015trust}. \cite{chow2018lyapunov,chow2019lyapunov} presents another class of method that formulates safe policy update under a Lyapunov constraint. \cite{perkins2002lyapunov} explored the relevance of Lyapunov functions in control and \citep{berkenkamp2017safe} used Lyapunov functions in RL to guarantee exploration such that the agent can return to a "region of attraction" in the model-based regime. In our work, we show that previous baselines rely on a backtracking recovery rule to ensure near constraint satisfaction and are sensitive to Q-function errors; we present a new method that uses a Lyapunov constraint with a barrier function to ensure a conservative policy update.

\textbf{Other notions of safety.} Recent works \citep{pham2018optlayer, dalal2018safe} use a safety layer along with the policy, which ensures that all the unsafe actions suggested by the policy are projected in the safe set. \cite{dalal2018safe} satisfies  state-based costs rather than trajectory-based costs. \cite{thananjeyan2020safety} utilizes demonstrations to ensure safety in the model based framework, and \cite{zhang2020cautious} learns the epistemic uncertainty of the environment by training the model in simulation for a distribution of environments, which is then used to cautiously adapt the policy while deploying on a  new test environment. Another line of work focuses on optimizing policies that minimize an agent's conditional value at risk (cVAR). cVAR~\citep{rockafellar2000optimization} is  commonly used in quantitative finance, which aims to maximize returns in the worst $\alpha\%$ of cases. This allows the agent to ensure that it learns safe policies for deployment that achieve high reward under the aleatoric uncertainty of the MDP \citep{tang2019worst,keramati2019being,tamar2014optimizing,kalashnikov2018qt, borkar2010risk, chow2014algorithms}.

\section{Conclusion}

In this work, we present a new method, LBPO, that formulates a safe policy update as an unconstrained policy optimization augmented by a barrier function derived from Lyapunov-based constraints. LBPO allows the agent to control the risk aversion of the RL agent and is empirically observed to be more robust to Q-function errors. We also present a simple baseline BACKTRACK to provide insight into previous methods' reliance on backtracking recovery behavior to achieve near constraint satisfaction. LBPO achieves fewer constraint violations, in most cases close to zero, on a number of challenging continuous control tasks and outperforms state-of-the-art safe RL baselines.

% \subsubsection*{Acknowledgments}
% Use unnumbered third level headings for the acknowledgments. All
% acknowledgments, including those to funding agencies, go at the end of the paper.

\newpage
\bibliography{iclr2021_conference}
\bibliographystyle{iclr2021_conference}

\appendix

\newpage
\section{Appendix}

\subsection{Safe policy update under the Lyapunov constraint}
\label{ap:lyapunov_derivation}

Let the safe initial (baseline) policy be given by $\pi_B$ and the Lyapunov function be defined as follows:
\begin{equation}
    \mathcal{L}_{\pi_B}(s_0,d_0) = \{L:S\rightarrow R_{\ge0}:\mathcal{B}_{\pi_B,c}[L](s)\le L(s),\forall s \in S; L(s_0)\le d_0 \}
\end{equation}
where c is the immediate cost function. Lyapunov functions depends on the safe baseline policy, an initial state and the cost constraint, and have the property that a one-step Bellman operator produces a value that is less that the value of the function at each state. We also know that the cost value function belongs to the set and hence the set is non-empty. Consider any Lyapunov function $L_{\pi_B}\in \mathcal{L}_{\pi_B}(s_0,d_0)$ and define:
\begin{equation}
    \mathcal{I}_{L_{\pi_B}} = \{\pi(.|s)\in\mathcal{P} : \mathcal{B}_{\pi,c}[L_{\pi_B}](s)\le L_{\pi_B}(s) \forall s\}
\end{equation}
to be set of policies consistent with the Lyapunov function $L_{\pi_B}$, called $L_{\pi_B}$-induced policies. These are the set of policies ($\pi\in\mathcal{I}_{L_{\pi_B}}$) for which a Bellman Operator $\mathcal{B}_{\pi,c}$ on a state $s$ produces a value that is less than the value of function at that state $L_{\pi_B}(s)$

Note that $\mathcal{B}_{\pi,c}$ is a contraction mapping, so we have
\begin{equation}
   V^c_{\pi}(s) = \lim_{k\rightarrow\infty} \mathcal{B}^{k}_{\pi,c}[L_{\pi_B}](s)\le L_{\pi_B}(s)~\forall s \in \mathcal{S}
\end{equation}
% \begin{equation}
%   D_{\pi_B}(s_0) = V^c_{\pi}(s_0)  \le L_{\pi_B}(s_0)\le d_0
% \end{equation}

From the definition of Lyapunov function, we also have that $L_{\pi_B}(s_0)\le d_0$. This implies that any policy induced by the Lyapunov function, i.e. policies in the $L_{\pi_B}$-induced policy set, are ``safe" i.e $V^c_\pi(s_0)=D_\pi(s_0)<d_0$. The method for safe reinforcement learning then searches for the highest performing policy within the safe policies defined by the set of $L_{\pi_B}$-induced policies. The objective here then is to design a Lyapunov function which contains the optimal policy, i.e optimal policy belongs to the set of $L_{\pi_B}$-induced policies so that the optimization restricted in this set indeed results in the solution of Eq. \ref{eq:CMDP}.

In general, the optimal policy $\pi^*$ does not belong the policies induced by the Lyapunov functions. \cite{chow2018lyapunov} show that without loss of optimality, the Lyapunov function that contains the optimal policy in its $L_{\pi_B}$-induced policy set can be expressed as $L_{\pi_B,\epsilon}(s) = \E{}{\sum_{t=0}^{\infty} \gamma^t (c(s_t)+\epsilon(s_t))|\pi_B, s}$, where $\epsilon(s_t)\ge0$.  The function $L_{\pi_B,\epsilon}(s)$  can be thought of as a cost-value function for policy $\pi_B$ augmented by an additional per-step cost $\epsilon(s_t)$. First, it can be verified that $\pi_B$ is indeed, in the set of $L_{\epsilon}$-induced policies:
\begin{equation}
\label{eq:epsilon_lyapunov_function}
    L_{\pi_B,\epsilon}(x) = \mathcal{B}_{\pi_B,c+\epsilon}[L_{\pi_B,\epsilon}](s) \ge \mathcal{B}_{\pi_B,c}[L_{\pi_B,\epsilon}](s)~~~\text{($\epsilon(s_t)>0~\forall s_t$)}.
\end{equation}
It was shown in ~\cite{chow2018lyapunov} that finding a state dependent function $\epsilon$ such that the the optimal policy is inside the corresponding $L_{\pi_B,\epsilon}$-induced set is generally not possible and requires knowing the optimal policy. As an approximation, they suggest to create the Lyapunov function with the largest auxiliary cost $\hat{\epsilon}$, such that $ L_{\pi_B,\hat{\epsilon}}(s) \ge \mathcal{B}_{\pi_B,c}[L_{\pi_B,\hat{\epsilon}}](s) $ and $L_{\pi_B, \hat{\epsilon}}(s_0)\le d_0$. The first condition is satisfied as shown in Eq.~\ref{eq:epsilon_lyapunov_function} when $\hat{\epsilon}(s)\ge0~\forall s$ and the second condition can be satisfied by the following derivation. Bold letters are used to denote vectors and $\boldsymbol{P^{\pi_B}_{s,s'}}$ is the transition probability matrix from state $s$ to $s'$ under policy $\pi_B$. The vectors contain the function value at each state. 
\begin{align}
    \boldsymbol{L_{\pi_B,\hat{\epsilon}}} &= \boldsymbol{d} +\boldsymbol{\hat{\epsilon}} + \gamma \boldsymbol{P^{\pi_B}_{s,s'}}\boldsymbol{L_{\pi_B,\hat{\epsilon}}}\\
    \boldsymbol{L_{\pi_B,\hat{\epsilon}}} &= (\boldsymbol{I}-\gamma \boldsymbol{P^{\pi_B}_{s,s'}})^{-1} \boldsymbol{d} + (\boldsymbol{I}-\gamma \boldsymbol{P^{\pi_B}_{s,s'}})^{-1} \boldsymbol{\hat{\epsilon}}\\
    \label{eq:finding_epsilon}
    \boldsymbol{1(s)^T}\boldsymbol{L}_{\pi_B,\hat{\epsilon}} &= \boldsymbol{1(s)^T} \boldsymbol{D}_{\pi_B} + \boldsymbol{1}(s)^T(\boldsymbol{I}-\gamma \boldsymbol{P^{\pi_B}_{s,s'}})^{-1} \boldsymbol{\hat{\epsilon}}
\end{align}
where $\boldsymbol{1(s)^T}$ is a one-hot vector in which the non-zero unit element is present at $s$. To ensure that the cumulative cost at the starting state is less than the constraint threshold, using Eq~\ref{eq:finding_epsilon} we have:
\begin{align*}
    L_{\pi_B,\hat{\epsilon}}(s_0)&\le d_0\\
     D_{\pi_B}(s_0) + \boldsymbol{1}(s_0)^T(I-\gamma \boldsymbol{P^{\pi_B}_{s,s'}})^{-1} \boldsymbol{\hat{\epsilon}}&\le d_0\\
\end{align*}
Notice that $\boldsymbol{1}(s_0)^T(I-\gamma \boldsymbol{P^{\pi_B}_{s,s'}})^{-1}\boldsymbol{1}(s)$ represents the total discounted visiting probability $\E{}{\sum_{t=0}^{\infty}\gamma^t\boldsymbol{1}(s_t=s)|s_0,\pi_B}$ of any state $s$ from the initial state $s_0$. Restricting $\hat{\epsilon}$ to be a constant function w.r.t state for simplicity, the value of $\hat{\epsilon}$ can be upper-bounded as:
\begin{equation}
    \hat{\epsilon}(s)\le (d_0-D_{\pi_B}(s_0)) / \E{}{\sum_{t=0}^\infty\gamma^t} = (1-\gamma)(d_0-D_{\pi_B}(s_0))
\end{equation}
% $\hat{\epsilon}(s)= d_0-D_{\pi_B}(s_0) / \E{}{\sum_{t=0}^\infty\gamma^T} = (1-\gamma)(d_0-D_{\pi_B}(s_0))$.\\
In summary, a Lyapunov function is obtained such that optimizing policies in the $L_{\pi_B,\hat{\epsilon}}$-induced set of policies, safety in ensured. For any policy $\pi$ to lie in the $L_{\pi_B,\hat{\epsilon}}$-induced set the following condition needs to hold $\forall~s\in\mathcal{S}$:
\begin{align*}
    L_{\pi_B,\epsilon}(s) &\ge T_{\pi,d} [L_{\pi_B,\epsilon}(s)]\\
    d(s)+\hat{\epsilon}(s)+\gamma\sum_{a}\pi_B(a|s)(\sum_{s'}P(s'|s,a)L_{\pi_B,\epsilon}(s')) &\ge d(s) + \gamma\sum_{a}\pi(a|s)(\sum_{s'}P(s'|s,a)L_{\pi_B,\epsilon}(s'))\\
\end{align*}
We can simplify further to get:
\begin{align*}
     \hat{\epsilon}(s)&\ge (\sum_{a}(\pi(a|s)-\pi_B(a|s))\left[\gamma\sum_{s'}P(s'|s,a)L_{\pi_B,\epsilon}(s')\right]\\
    \hat{\epsilon}(s)&\ge (\sum_{a}(\pi(a|s)-\pi_B(a|s))\left[\gamma\sum_{s'}P(s'|s,a)L_{\pi_B,\epsilon}(s') + d(s)+\hat{\epsilon}(s)\right]\\
    \hat{\epsilon}(s)&\ge \left[\sum_{a}(\pi(a|s)-\pi_B(a|s))Q_{L_{\pi_B,\epsilon}}(s,a)\right]
\end{align*}

where
\begin{equation}
    \label{eq:ap_lyapunov_q}
    Q_{L_{\pi_B,\epsilon}}(s,a)=d(s)+\hat{\epsilon}(s)+\gamma \sum_{s'}P(s'|s,a)L^{\pi_B,\epsilon}_{\hat{\epsilon}}(s')
\end{equation}
This can be extended to continuous action spaces to get the following objective:
\begin{equation}
    \label{eq:ap_lyapunov_constraint}
    \pi_+(.|s) =  \max_{\pi\in\mathcal{P}} J_{\pi}(s_0),~s.t~\int_{a\in \mathcal{A}}(\pi(a|s)-\pi_B(a|s)) Q_{L_{\pi_B,\hat{\epsilon}}}(s,a) da\le \hat{\epsilon}(s)~\forall s \in \mathcal{S}
\end{equation}
Using the Lyapunov function, the trajectory-based constraints of CMDP are converted to a per-state constraint (Eq.\ref{eq:ap_lyapunov_constraint}), which are often much easier to deal with.

In the case of deterministic policy, the policy update becomes:

\begin{equation}
    \pi_+(.|s) =  \max_{\pi\in\mathcal{P}} J_{\pi}(s_0),~\text{s.t}~ Q_{L_{\pi_B,\hat{\epsilon}}}(s,\pi(s))-Q_{L_{\pi_B},\hat{\epsilon}}(s,\pi_B(s)) \le \hat{\epsilon}(s)~\forall s \in \mathcal{S}
\end{equation}

An intuitive way to understand the constraint in deterministic policies is to see that at every timestep we are willing to tolerate an additional constant cost of $\epsilon$ compared to the baseline safe policy. At the start state, the maximum increase in expected cost will be $\sum_{t=0}^\infty\gamma^t\epsilon=\frac{\epsilon}{1-\gamma}$. We want that the new expected cost by less than the threshold, i.e $D_{\pi}(s_0)+\frac{\epsilon}{1-\gamma}\le d_0$ which gives us the Lyapunov constraint equation.

\subsubsection{From Lyapunov Functions to cost Q Functions}
\label{ap:lyapunov_to_costq}
Using the definition of $Q_{L_{\pi_B,\hat{\epsilon}}}(s,a)$ from Eq.~\ref{eq:lyapunov_q} and when $\hat{\epsilon}(s)$ is a constant function (denote by $\hat{\epsilon}$), we can replace $Q_{L_{\pi_B,\hat{\epsilon}}}$ by $Q^{C}_{\pi_B}$,
\begin{align*}
  Q_{L_{\pi_B,\hat{\epsilon}}}(s,a)&=c(s)+\hat{\epsilon}+\gamma \sum_{s'}P(s'|s,a)L_{\pi_B,\hat{\epsilon}}(s')\\
  &= c(s)+\hat{\epsilon}+\left[\gamma \sum_{s'}P(s'|s,a)[c(s')+\hat{\epsilon}+\sum_{s''}P^{\pi_B}_{(s''|s')}(L_{\pi_B,\hat{\epsilon}}(s''))]\right]\\
  &= \sum_{t=0}^{\infty}\gamma^t\hat{\epsilon} +  \E{}{\sum_{t=0}^{\infty}\gamma^t c(s_t)|\pi_B,a_0=a, s_0= s}\\
&= \sum_{t=0}^{\infty}\gamma^t\hat{\epsilon} +  Q^{C}_{\pi_B}(s,a)
\end{align*}
which is the cost Q function, since the Lyapunov function $Q_{L_{\pi_B,\hat{\epsilon}}}(s,a)$ and the cost-Q function $Q^{C}_{\pi_B}(s,a)$ only differ by a constant ($\sum_{t=0}^\infty \gamma^t\hat{\epsilon}$).

% \newpage

\subsection{Additional Results}

\subsubsection{Benchmarks on OpenAI safety gym}
\label{ap:benchmarks}

In this section, we present the training curves for all the OpenAI safety gym environments with Point and Car robot. Figure~\ref{ap:benchmarks} shows the Average Cost and Average return for these environments. The dotted red line indicates the constraint threshold which is kept to be 25 across all environments. We observe that LBPO rarely violates constraint during training. Table~\ref{tab:mujoco_all_reward_unnormalized} shows the raw cumulative returns of the converged policy for different methods on the safety environments. We average all results over 3 random seeds.

% Table~\ref{tab:mujoco_all_constraint} show the fraction of unsafe behavior policies encountered during training for 2e7 timesteps for all the baselines - PPO, PPO-lagrangian, CPO, SDDPG, BACKTRACK and compare it to LBPO. Table~\ref{tab:mujoco_all_reward} shows the performance of the policy at convergence for all the safety methods. It is observed that LBPO trades off little performance to more accurate constraint satisfaction. For these results, we run experiments till convergence for 2e7 timesteps, and the policy is updated every 30000 timesteps. 
We observe that in tasks with Doggo robot, none of the methods are able to obtain good performing policy. We attribute this to be the difficulty of Doggo environments, involving an inherent tradeoff of reward with cost. In the environment PointGoal2, we are unable to obtain safe policies even when training an RL agent solely on the cost objective. LBPO still outperforms baselines for constraint satisfaction on this environment.
\begin{figure}[h]
    \centering
    \includegraphics[width=\linewidth]{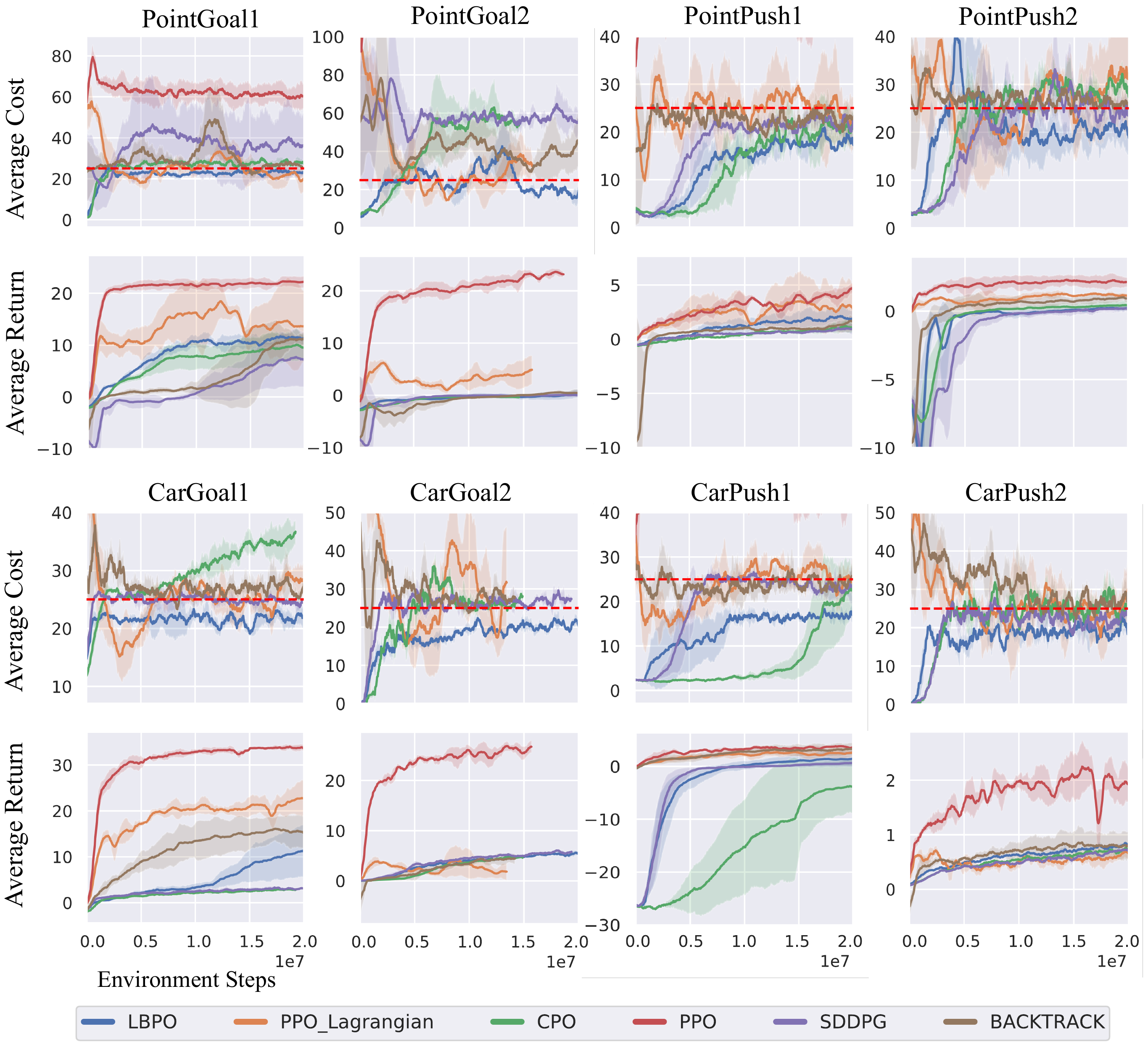}
    \caption{Training curved for LBPO in comparison to baselines: PPO, PPO-lagrangian, CPO, SDDPG. We also compare against our simple baseline BACKTRACK here. For each environment, the top row shows the Average undiscounted cumulative cost during training, and bottom row shows the Average undiscounted return. PPO often has large constraint violations and is clipped from some plots, when its constraint violations are high. Red dashed line in Average Cost plots shows the constraint limit which is 25 in all environments.}
    \label{fig:lbpo_benchmark}
\end{figure}

\begin{table}[h]
    \vspace{5mm}
    \centering
    \footnotesize
    % \resizebox{\textwidth}{!}{%
    \begin{tabular}{c|c|c|c|c|c|c}
        \toprule
         Method & PPO & PPO-lagrangian & CPO & SDDPG& BACKTRACK & LBPO\\
        \midrule
        PointGoal1 &22.99&\textbf{19.00}&10.26&10.45&15.54&11.06\\
        PointGoal2 &23.04&\textbf{4.60}&-0.37&-0.08&1.04&0.61\\
        PointPush1 & 4.61 & 3.04 & 1.73 & 2.71 & 2.43 & \textbf{3.15} \\
        PointPush2 & 2.15 & \textbf{1.04} & 0.46 & 0.48 & 0.89 & 0.77 \\
        CarGoal1 & 34.62 & 15.55 & 2.76 & 3.38 & \textbf{17.22} & 13.03 \\
        CarGoal2 & 26.70 & 1.78 & 4.60 & \textbf{5.74} & 4.35 & 5.62 \\
        CarPush1 & 3.89 & 2.72 & -3.13 & 1.69 & \textbf{3.38} & 1.89 \\
        CarPush2 & 2.03 & 0.72 & 0.82 & 0.75 & 0.81 & \textbf{0.94} \\
        DoggoGoal1 &38.76 & -0.65 & 0.14 & 0.10 & 0.15 & \textbf{0.28} \\
        DoggoGoal2 &18.38 & \textbf{0.31} & 0.04 & 0.06 & 0.06 & 0.06 \\
        DoggoPush1 & 0.82 & \textbf{0.07} & 0.01 & 0.00 & 0.06 & 0.01 \\
        DoggoPush2 & 1.10 & \textbf{0.08} & -0.00 & -0.00 & 0.07 & -0.01 \\
        \bottomrule
    \end{tabular}
    % }
    \caption{Cumulative unnormalized return of the converged policy for each safety algorithm. LBPO   tradeoffs return for better constraint satisfaction. Bold numbers show the best performance obtained by a safety algorithm (thus excluding PPO).}
    \label{tab:mujoco_all_reward_unnormalized}
\end{table}

\newpage
\subsubsection{Backtracks in CPO and SDDPG}
\label{ap:backtracks}
\begin{figure}[h]
    \centering
    \includegraphics[width=0.7\linewidth]{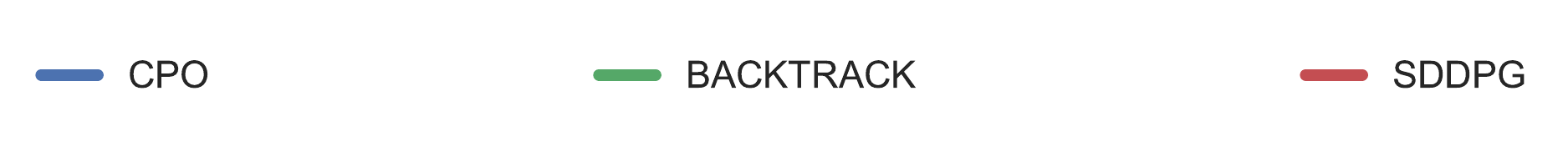}
    \includegraphics[width=\linewidth]{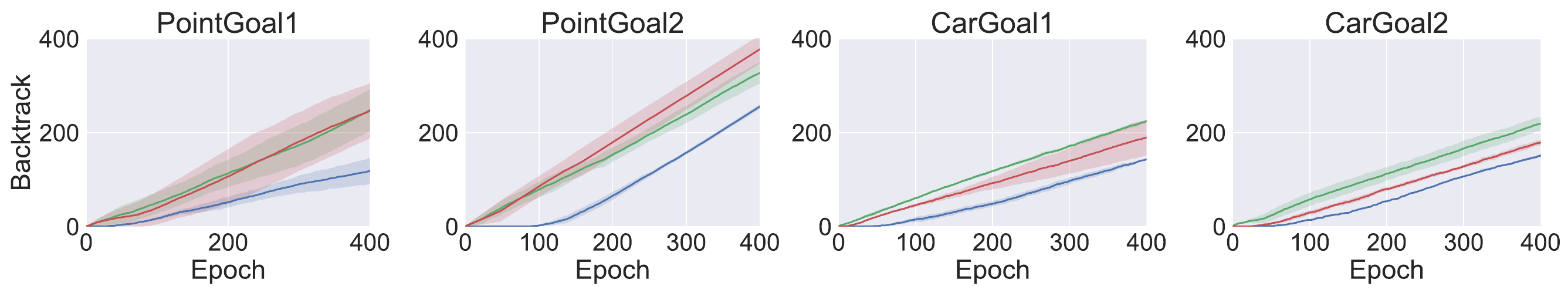}
    \caption{We compare the cumulative number of backtracking steps taken by CPO and SDDPG to BACKTRACK method for the first 400 epochs/policy updates.}
    \label{fig:backtrack_analysis}
\end{figure}
Figure~\ref{fig:backtrack_analysis} shows the cumulative number of backtracks performed by each method CPO, SDDPG, BACKTRACK during the first 400 policy update steps. We see the CPO and SDDPG performs a high number of backtracks, often comparable to the method BACKTRACK which relies explicitly on backtracking for safety.

\subsection{Implementation Details}
\label{ap:implementation}
In LBPO, Q-functions (both reward and cost) have network architecture comprising of two hidden layers of 64-hidden size each. The policy is also a multilayer neural network comprising of three hidden layers of 256 units each. LBPO policies are deterministic and have a fixed exploration noise in the action space given by $\mathcal{N}(0,0.05)$. Our trust region update for the policy takes into account the exploration noise which makes our behavior deployment policy stochastic.  We use $N=30$ trajectories each of 1000 horizon length for generating our on-policy samples. These samples are used for estimation of $\hat{\epsilon}$ and evaluating the Q functions. We update the policy under a trust region followed by a line search with exponential decay which ensures that the resulting update is indeed satisfying the KL constraint as well the safety Lyapunov constraint. We do a hyperparameter search for $\beta$ in the set [0.005, 0.008, 0.01, 0.02] to find the best tradeoff between cost and reward and observe that a value of 0.005 works well across most environments. PointGoal1, PointPush1, PointPush2, CarGoal1, CarPush1, CarPush2, DoggoGoal1, DoggoPush1, DoggoPush2 use beta value of 0.005. CarGoal2 and DoggoGoal2 uses value of 0.008 and Pointgoal2 uses beta value of 0.01. We ignore the barrier loss if $\beta$ is sufficiently low. We call this parameter $\beta$-thres and it is set to 0.05 across all environments. 
\\
\begin{algorithm}
\vspace{2mm}
 Initialize parameterized actor $\pi_{\phi}$ with a safe initial policy, reward Q-function $Q^R_{\theta}$ and a cost Q function $Q^C_{\theta}$
 \begin{algorithmic}[1]
 \ForEach{$i \leftarrow \ 1$... $\text{Iter}$}
    \State \textbf{Step 1: }Collect $N$ trajectories $\{\tau\}_{j=1}^N$ using the safe policy $\pi_{\phi,i-1}$ from previous iteration $i-1$.
    \State \textbf{Step 2: }Using the on-policy trajectories, evaluate the reward Q-function and the cost-function, by minimizing the respective bellman residual of the TD-($\lambda$) estimate.
    \State \textbf{Step 3: }Update the policy parameters by minimizing the objective in Eq \ref{eq:LBPO_objective}. 
    \begin{center}
        $\text{min}_{\phi}~ \E{s\sim\mathcal{R}}{-Q^R_{\pi_{\phi,i-1}}(s,\pi_{\phi}(s)) +  \psi(Q^{C}_{\pi_{\phi,i-1}}(s,\pi_{\phi}(s)))}$\\
        $~~~~~~~~~~\text{s.t}~D_{\text{KL}}(\pi_\phi+\mathcal{N}(0,\delta)\|\pi_{\phi,i-1}+\mathcal{N}(0,\delta))< \mu$
    \end{center}
    \State \textbf{Step 4: } Set $\pi_{\phi,i}$ to be the safe policy resulting from the update in Step 3 ($\pi_\phi$).
    \EndForEach
    \end{algorithmic}
    \caption{LBPO}
 \label{algo:LBPO}
 \vspace{2mm}
\end{algorithm}

We obtain safe initial policies for benchmarking by pretraining the policy using standard RL methods to minimize the cumulative cost. Although this strategy is usually not suitable for deployment in real-world as the pretraining might itself violate safety constraints, we can use simple hand-designed safe policy for initializing the method in real-world experiments.

To ensure fair comparison across methods, we use the same safe initial policy for each of the safety methods. Note that, our results for CPO~\citep{achiam2017constrained} significantly differ from the benchmarks shown in ~\citep{Ray2019}  due to the fact that we initialize CPO from safe policy contrary to their approach. We also keep the same policy architecture across methods although CPO, PPO and PPO-lagrangian uses policies with learned variance so as to replicate the original behavior of these methods.

\begin{table}[h]
  \begin{center}
    \caption{LBPO Hyperparameters}
    \label{tab:table1}
    \begin{tabular}{l|l}
      \toprule % <-- Toprule here
      \textbf{Hyperparamater} & \textbf{Value}\\
      \midrule % <-- Midrule here
      N & 30\\
      $\beta$ & 0.005\footnotemark\\
      $\beta$-thres  & 0.05\\
      Policy learning rate & 3e-4\\
      Q-function learning rate  & 1e-3\\
      Trust region ($\mu$)  & 0.012\\
      $\lambda$ & 0.97\\
      $\delta$ & 0.05\\
      Horizon & 1000\\
      \bottomrule % <-- Bottomrule here
    \end{tabular}
  \end{center}
\end{table}
\footnotetext[1]{$\beta$ is set to 0.005 for most environments. Appendix~\ref{ap:implementation} describes specific value of $\beta$ for each environment.}
We implement our version of SDDPG which uses the $\alpha$-projection technique as shown in~\citep{chow2019lyapunov}. A brief discussion of practical issue faced in the implementation is present in Section~\ref{sec:experiments}. We use behavior cloning to distill the policy with the projection layer into a parameterized multilayer perceptron policy. We run 100 iterations of behavior cloning with learning rate of 0.001. We implement a line search with exponential decay in parameter space to ensure that the resulting update do not violate the Lyapunov constraints to incorporate additional safety. We use similar policy architecture as LBPO for $\alpha$-SDDPG.

% Since our policy is gaussian with fixed standard deviation, we can rewrite the expected KL constraint as a expected L2 constraint between the deterministic actions.

% \section{Appendix}
% You may include other additional sections here.

\end{document}